Viewpoint

# When Clinical Judgment Becomes a Predictive Shortcut: Two Questions Clinical AI Must Answer Before Deployment

Jing Yang[1], Long R. Jiao[2], Zongjiu Zhang[1*], Xiujun Cai[3*]

[1]School of Biomedical Engineering, Tsinghua University, Beijing 100084, China

[2]Department of Surgery and Cancer, Imperial College London, London W2 1NY, United Kingdom

[3]Sir Run Run Shaw Hospital, Zhejiang University School of Medicine, Hangzhou 310016, China

* Corresponding authors: Xiujun Cai (srrsh_cxj@zju.edu.cn) and Zongjiu Zhang (zhangzongjiu@tsinghua.edu.cn)

## Abstract

Clinical artificial intelligence can predict deterioration, complications, and other outcomes using tests, treatments, and documentation that reflect clinicians' existing judgments. Performance based on this information may nevertheless be interpreted as evidence that a model can inform judgment before clinicians recognize a change. Extrapolating that performance to use before clinical recognition, or to hospitals with different care processes, may overestimate prediction when the relevant information is absent.

We propose an evaluation framework that matches evidence to capability claims. It first specifies the intended clinical stage, examines whether key inputs depend on existing clinical judgment and are available at that stage, and evaluates performance for a defined outcome and prediction horizon under those information conditions. For use before clinical recognition, prespecified observable measures, such as the first documented suspicion or an order to investigate a clinical hypothesis, serve as recognition proxies; plausible alternatives support sensitivity analyses.

A stronger claim that a model has recognized a departure from the clinical course before a declared recognition proxy requires an as-of state reference, target-matched performance evaluation, and temporal validation. References use only information available at each time, with criteria informed by clinical standards, patient baseline, and stage of care. Depending on the target, validated rules or structured review may be used. Manual review can be restricted to a prespecified validation sample, with independent reassessment to evaluate reproducibility.

The framework helps developers, clinicians, and hospital leaders identify the clinical stages and information prerequisites supported by performance evidence. It guards against extrapolating predictive advantages supplied by existing care processes to use before recognition or at other hospitals. Before deployment, two questions must be answered: does the information underpinning performance exist when support is most needed, and does adequate validation support performance when those conditions change? These questions remain a blind spot of current evaluation practice: leakage checks, overall predictive performance, and external validation neither address nor answer them. We establish these questions as deployment prerequisites and provide an operational approach—information-condition checks, an as-of state reference, and proxy-relative timing tests—for answering them directly.

## From predictive performance to expectations for clinical use

Hospitals adopt clinical AI for reliable judgment, but development performance need not transfer to the intended hospital. For an early Epic Sepsis Model, the developer reported an area under the receiver operating characteristic curve (AUROC) of 0.76–0.83[1]; Wong and colleagues reported an encounter-level AUROC of 0.63 in external validation across 38,455 hospitalizations[1]. This cross-study difference underscores the need for setting-specific validation. Performance must also support the intended clinical stage. Electronic health records encode both patient state and clinical judgments and responses. The latter may predict outcomes but requires separate evaluation for use before clinical recognition.

Shortcut learning provides a computational perspective[2]. Supervised clinical prediction models commonly minimize mean training prediction loss through empirical risk minimization (ERM). ERM does not distinguish patient-state clues from existing clinical judgment; models may perform well using judgment encoded in care providers' actions. Under studied data and optimization conditions, theory and experiments show gradient-trained neural networks favoring simpler features, sometimes ignoring more predictive complex features[3]. Clinical actions can encode integrated real-time judgments, including experience-based recognition poorly captured in structured records, and carry substantial predictive information. Across approximately 42.9 million hospitalizations, models using only administrative and care-team activity data approached published full electronic health record model benchmarks without laboratory results, vital signs, or clinical notes[4]. For an early Epic Sepsis Model, AUROC was 0.62 using all predictions before sepsis criteria were met, but 0.47 after excluding predictions following the first evidence of treatment planning[5]. Overall performance incorporating clinical responses cannot directly represent performance without that information.

An updated Epic Sepsis Model illustrates why performance before recognition must be evaluated for the specific model, proxy, and window, rather than inferred from overall performance or an earlier model. Encounter-level AUROC changed from 0.82–0.92 to 0.80–0.90 when evaluation was bounded by a prespecified recognition proxy preceding the time Sepsis-3 criteria were met. The proxy was the earliest antibiotic, lactate, or culture order; where none preceded Sepsis-3 criteria, the Sepsis-3 time remained the reference[6].

This evaluation framework helps developers, clinicians, and hospital leaders identify which clinical stages and information prerequisites the evidence supports. Care-process predictive advantages should not be extrapolated to use before recognition or at other hospitals. The framework asks whether necessary information exists when support is most needed and whether validation supports performance when conditions change. Formal arguments and counterexamples show that improved event prediction can coexist with model signals occurring after recognition proxies, without conventional temporal leakage (Multimedia Appendix 1, Propositions S1–S5).

## How clinical judgment enters inputs and evaluation targets

Data leakage generally involves training or evaluation using information unavailable at prediction, inflating reported performance[7]. In records, temporal or availability leakage includes later-written notes, unreported results, and discharge codes used for earlier predictions. Our concern is distinct: an order prompted by suspected deterioration may already be visible when a model is queried. Its use is temporally valid, but the encoded judgment cannot be assumed to exist before recognition. Leakage checks establish necessary temporal compliance; they do not establish suitability for use before recognition. This dependence requires separate evaluation.

Existing judgment creates dependence through inputs and through endpoint definition and recording (Table 1). Records reflect both illness and care. Changes may prompt suspicion, monitoring, investigations, treatment, and documentation[8,9]. Event definitions and times may therefore reflect patient-state changes, suspicion, clinical responses, and formal recording[10,11].

On the input side, observation frequency, unplanned tests, and consultations may be available at query time and improve prediction[4]. They may also indicate that the team is already responding. Such performance does not establish recognition before those actions; for that intended use, dependence on these clues risks becoming a predictive shortcut (Fig. 1a).

On the target side, judgment may shape endpoint definition, confirmation, and recording. Outcomes based on initiated investigations, clinical grading, or documentation rules remain legitimate prediction targets, but their meaning and timing require interpretation against those processes. They cannot directly replace an independent reference for the current departure and when it first became recognizable (Table 1). Removing obvious action features therefore does not remove the need to examine target construction.

Truncating predictions at first treatment-planning evidence evaluates performance before that proxy; earlier undocumented suspicion may still influence inputs. Requirements depend on the claim: future-outcome prediction requires evaluation at the intended stage; recognition of a departure already meeting prespecified recognizability criteria additionally requires an as-of state reference and temporal comparison with recognition proxies. Table 1 distinguishes four issues. Temporal or availability leakage can be checked by reconstructing and freezing information visible at query time. The other three can persist despite valid timestamps and motivate this framework.

Table 1. Four issues requiring distinct treatment in prediction from clinical records.

| **Source of concern** | **Typical clues or targets** | **Evaluation issue** | **Temporal or availability leakage?** | **Required response** |
|---|---|---|---|---|
| I. Temporal noncompliance: temporal or availability leakage (established in machine learning) | | | | |
| Later-available information enters earlier predictions | Discharge codes, later-written notes, results not yet reported | Temporal or availability leakage | Yes | Freeze the information actually visible at query time |
| II. Information dependence despite temporal compliance: dependence on existing clinical judgment (the focus here) | | | | |
| Existing clinical judgment enters inputs through clinical actions | Observation frequency, unplanned lactate testing, imaging, consultations | Care-process signals can improve prediction; the capability they support requires separate definition | No; the information may already be available | Trace input origins; evaluate performance at the intended stage and before declared recognition proxies; test robustness with alternative proxies |
| Clinical judgment contributes to endpoint definition, confirmation, or recording | Complication grades assigned at fixed dates; first antibiotic time | The evaluation target is intertwined with care processes | No; the issue concerns target construction | Specify the endpoint's clinical meaning and confirmation process; claims of recognizing current changes require a separate as-of state reference |
| Treatment after recognition changes | Mortality following early antibiotics | Treatment affects the predicted | No; this is a causal issue | Specify prediction under observed care; studies of action effects |

| Source of concern | Typical clues or targets | Evaluation issue | Temporal or availability leakage? | Required response |
|---|---|---|---|---|
| outcomes | | outcome[12,13] | | require an appropriate causal or clinical comparative design |

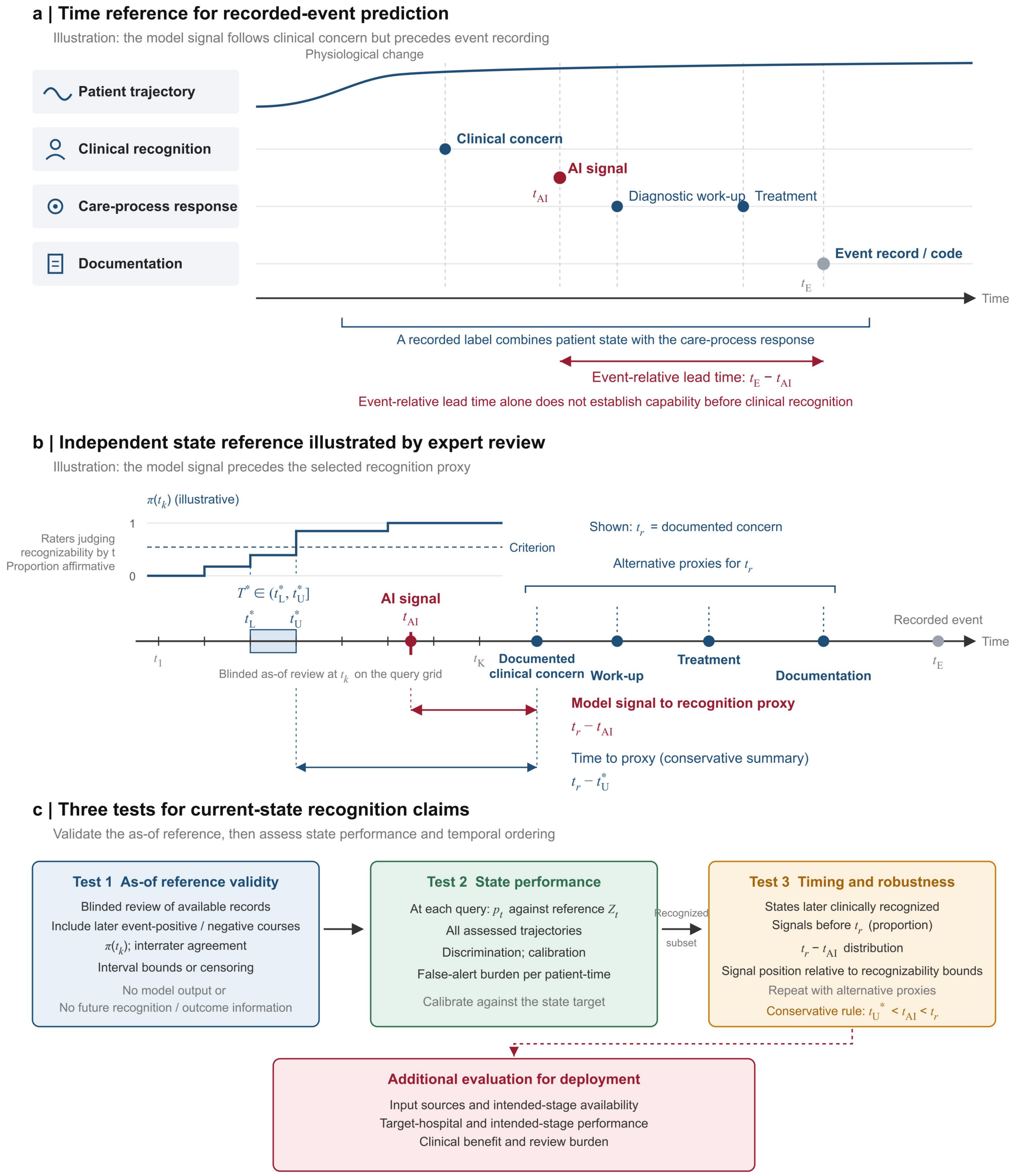

Fig. 1. Detailed validation and temporal references for current-state recognition claims.
(a) Recorded-event evaluation. State changes may prompt suspicion, tests, treatment, and event recording. Responses can supply inputs and affect endpoint confirmation and timing; event-based lead time may include time after recognition begins.
(b) Independent as-of reference, illustrated by expert review. Assess departure against prespecified criteria using contemporaneous evidence; define recognition separately by proxy. This illustrates detailed state-recognition validation, not a fixed procedure for all future-outcome prediction studies. Interpret signals before the proxy using the recognizability interval. At $t_1 \ldots t_k$, blinded raters assess recognizability; $\pi(t)$ is the affirmative proportion. $t^*_U$ first meets criteria; $t^*_L$ is the last preceding negative. If the adjudicated interval is valid, $T^* \in (t^*_L, t^*_U]$. Compare model crossing $t_{AI}$ with proxy $t_r$; alternatives change $t_r$ and interval $t_r - t_{AI}$. Conservatively, signals no later than $t^*_L$ are anticipatory; those in $(t^*_L, t^*_U]$ are temporally indeterminate; signals after $t^*_U$ but before $t_r$ provide sufficient temporal evidence for state recognition before the proxy, with lead time $t_r - t_{AI}$.

(c) Three tests, illustrated by panel b. Test 1 uses prespecified independent assessment, intermixing trajectories with and without subsequent recorded events; report $\pi(t)$, agreement, $(t^*_L, t^*_U]$, and left/right censoring, without model outputs or $t_r$. Against $Z_t$, Test 2 evaluates $p_t$ discrimination, calibration, and false-alert burden per patient-time at a prespecified threshold across all assessed trajectories. Test 3 reports first-crossing proportions before $t_r$, interval $t_r - t_{AI}$, and $t_{AI}$ relative to $(t^*_L, t^*_U]$ among subsequently recognized transitions; repeat with alternative proxies. Lower box: information conditions and setting-specific performance for all uses; action benefits and burdens require separate evaluation. Red: model signal; blue: proxy; gray: recorded event.

## Evaluating prediction for the intended use

### Specify the clinical stage and prediction task

Specify patients, stage of use, outcome, horizon (for example, 24 hours), and update frequency. We call the prediction time the query time. Prediction before relevant clinical judgment and risk stratification after investigations may require different information and support different uses. Subsequent outcomes may supply labels, never earlier inputs; report outcome ascertainment, follow-up, and treatment conditions.

### Examine information conditions for inputs and endpoints

Distinguish collection, result-generation, and clinical-availability times for key inputs, and determine whether they arose from routine monitoring, suspicion, or established care protocols. Removing orders alone is insufficient: suspicion may also prompt physiological measurement. Explain how investigations, interventions, and documentation determine case ascertainment and event timing. Availability before query time establishes temporal compliance; suitability for the intended stage also depends on how information was generated.

### Validate within windows matching the intended use

For use before recognition, prespecify observable proxies, such as first documented suspicion or relevant test orders, and evaluate future-outcome prediction in eligible patients and queries preceding them. Specify handling of absent proxies, routine screening orders, and missing timestamps; repeat analyses with plausible alternatives. Findings support prediction relative to the declared proxy, not necessarily precedence over clinicians' actual awareness.

Report discrimination, calibration, and uncertainty, specifying population, queries, and horizon. Proxy-based truncation changes these distributions and outcome prevalence; performance differences therefore cannot directly quantify clinical judgment's contribution. Use comparable patients and horizons where possible, and separately report performance at the intended stage. Interval estimates must account for within-patient query dependence. Contributions of particular clues require explicit provenance and comparison designs; feature importance or a single ablation cannot establish judgment's causal contribution.

Future-outcome prediction is evaluated primarily using continuous risk scores. Thresholds, event capture, false alerts, and review burden become necessary for specific signal or action claims. These requirements connect intended use, information conditions, and interpretation of judgment-dependent performance, complementing TRIPOD+AI and DECIDE-AI reporting guidance[14,15]. Table 3 summarizes supported interpretations and deployment implications.

### Detailed validation of claims to recognize current clinical changes

Claims of recognizing a clinically meaningful change from contemporaneous information before a recognition proxy require an independent as-of state reference. We illustrate this using departures from the clinical course. Defining the state as “not yet clinically recognized” would make temporal precedence definitional rather than independently testable.

The target is a clinically meaningful change assessed against prespecified criteria informed by clinical standards, patient baseline, and stage of care, termed a departure from the clinical course. Its first recognizable time is $T*$. “Recognizable” means available evidence suffices for assessment, whether or not the model or clinical team has recognized it.

Recognition timing is compared only after defining the target. If $T* < t_r$, the departure is recognizable before the proxy; $t_r$ is defined by a prespecified observable measure. First recognizability depends on available evidence, not biological onset. Biological and clinical stages of sepsis can be asynchronous[16], and records cannot precisely reconstruct biological onset (Fig. 1b).

Assessment criteria. Prespecify the change, clinical basis, and operational criteria, such as sustained baseline-relative abnormalities or corroborating evidence. Clinical standards, baseline, and stage inform assessment; expected recovery is one implementation, without requiring a complete future trajectory for every patient. Criteria must precede evaluation, cannot be validated by the evaluated model or the patient's subsequent recognition, care, or outcome, and require reported uncertainty.

For example, reviewers assessing postoperative day 3 after pancreaticoduodenectomy see only observations, drainage volume and characteristics, and laboratory results available by 06:00. Records from patients with and without subsequent complications are intermixed. Reviewers assess departure from expected recovery and confidence without knowing subsequent fistula grades. Self-resolving departures still count as state transitions: departure and subsequent outcome are distinct targets.

Information boundary. Let $\mathcal{F}_t$ denote the σ-algebra generated by record information available by query time t, bounding information for both model and reference. Inputs and assessment must be measurable with respect to $\mathcal{F}_t$, excluding unreported results, later-written notes, and discharge codes. Final database values cannot stand in for earlier visible values.

Information arrival and physiological change have distinct timing constraints. Collection, result generation, and visibility may be hours apart[17]; in acute kidney injury, creatinine can lag filtration decline by hours to days[18]. Reconstructing contemporaneous records does not itself establish that evidence sufficed to recognize departure.

Recognition proxies. Clinicians' first awareness is difficult to observe directly. Prespecify observable measures defining $t_r$, such as documented suspicion, orders investigating a clinical hypothesis, or treatment initiation (Table 2). Proxies provide auditable time references; their absence does not establish absence of clinical judgment.

Proxy choice changes $t_r$. Consensus definitions of pancreatic fistula and posthepatectomy liver failure involve fixed postoperative dates[19,20], so documentation-defined $t_r$ may follow initial concern by days. Interpret precedence relative to the proxy and test plausible alternatives.

Table 2. Clinical recognition proxies and their influence on time reference $t_r$.

| Recognition proxy | Examples (general clinical care; perioperative care) | Strengths | Interpretive limitations | Influence on $t_r$ |
|---|---|---|---|---|
| Clinical suspicion | Relevant clinical notes, problem-list entries, documented | Closer to the cognitive act of | Inconsistent documentation; delayed | Potentially early; depends on |

| Recognition proxy | Examples (general clinical care; perioperative care) | Strengths | Interpretive limitations | Influence on $t_r$ |
|---|---|---|---|---|
| | concerns;<br>"Suspected bile leak" or "monitor drain output" in ward-round notes | recognition | entry; noisy natural-language extraction | documentation completeness and timeliness |
| Diagnostic investigation and assessment | Blood cultures, lactate, imaging, specialist consultation;<br>unplanned drain-fluid amylase or bilirubin tests, abdominal CT, surgical or ICU assessment | Clearer timestamps; reflects ongoing investigation | May reflect protocolized screening rather than suspicion | May precede treatment; depends on routine screening protocols |
| Treatment | Antibiotics, fluids, vasopressors, ICU transfer;<br>reoperation, interventional drainage, escalation of care | Relatively well defined; reflects action taken | Usually beyond the early window; already reflects a clinical decision | Usually later; reflects a treatment decision |
| Documentation | Diagnostic codes, complication registries, discharge diagnoses;<br>ISGPS/ISGLS or Clavien–Dindo grading records | More standardized; easier extraction | May be assigned retrospectively after outcome resolution; poorly represents first recognition | Usually later; may lag by days |

Note: Row order does not imply actual ordering in individual patients. Prespecify extraction rules for $t_r$, using the first qualifying timestamp that was actually visible. Explain handling of routine screening orders, absent proxies, and missing timestamps; report alternative-proxy analyses. CT, computed tomography; ICU, intensive care unit; ISGPS, International Study Group on Pancreatic Surgery; ISGLS, International Study Group of Liver Surgery.

Assessment and intervals. Prespecify query grid $t_1 < \ldots < t_K$. At each $t_k$, raters or rules assess recognizability using only records available by $t_k$; $C(t_k)$ indicates criteria fulfillment. Use clinically justified, validated rules or structured clinical review. Multiple-rater schemes may prespecify aggregation by majority, confidence, and evidence requirements.

The target records whether departure has ever become recognizable. Once reached, this historical state remains positive despite recovery; it does not imply continuing abnormality.

Discrete queries interval-censor $T^*$. The first query satisfying $C(t_k)$ is $t_U^*$; the last preceding negative query is $t_L^*$. If assessment validly bounds recognizability onset, $T^* \in (t_L^*, t_U^*]$. Conceptually, $T^* < t_r$ denotes recognizability before the proxy; $t_U^* < t_r$ is a conservative sufficient condition established from the valid interval. If $t_r$ falls within $(t_L^*, t_U^*]$, strict ordering remains uncertain.

A positive initial truncated record left-censors $T^*$. If all valid queries before $t_r$ are negative, report right censoring at the last valid negative query; do not extend negativity across the remaining window.

Historical target monotonicity differs from observed assessment monotonicity. With multiple raters, retain raw affirmative proportions $\pi(t_k)$ and agreement; for rules, report validation and indeterminate results. Prespecify stricter criteria or monotonic constraints and retain unconstrained results.

Continuous scores and threshold crossings. Let $Z_t = 1\{T^* \leq t\}$ indicate recognizability reached by t. Actual inputs $X_t$ satisfy $\sigma(X_t) \subseteq \mathcal{F}_t$; reference assessment may use all contemporaneously available records. Denote the true conditional probability by $p_t^* = P(Z_t = 1 \mid X_t)$ and its model estimate on the prespecified grid by $p_t = f(X_t)$. At declared threshold c, first crossing is $t_{AI} = inf\{t : p_t \geq c\}$. Evaluate actual outputs for discrimination and calibration. Here $p_t$ concerns historical recognizability, distinct from future-outcome risk. Outcome scores may be tested for state discrimination; state-probability interpretation requires target-specific re-estimation or calibration and independent validation. No crossing means no alert. First crossing is a threshold-specific system time, not the first emergence of predictive information.

Recognition delay. Relative to proxy time $t_r$, delay from first recognizability lies between $t_r - t_U^*$ and $t_r - t_L^*$. A conservative single-value summary is $t_r - t_U^*$, excluding time of uncertain recognizability.

Model lead time. For signals before $t_r$, $t_r - t_{AI}$ is simply the crossing-to-proxy interval. After validating the target and performance, locate $t_{AI}$ relative to $(t_L^*, t_U^*]$ to determine whether the interval represents recognizable departure before the recognition proxy (Table 3; Multimedia Appendix 1, Propositions S5 and S8).

Conventional warning systems target subsequent arrest, ICU transfer, death, or bundled treatment[21,22]; physiological antecedents are documented[23,24]. Claims of recognizing departure before clinical recognition additionally require the target and temporal evidence in Table 3.

Change-point detection, onset estimation, informative-observation, and latent-state methods can support modeling[9,11,25]. Clinical interpretation still requires an explicit reference target and temporal evidence matching the intended stage.

Table 3. Interpretations and deployment implications supported by evaluation evidence.

| Observed evidence | Supported interpretation | Deployment implications and additional evidence |
|---|---|---|
| High performance for recorded events | Predicts the recorded outcome under evaluated conditions | Use before proxy $t_r$ requires direct evaluation at that stage; cross-site use additionally requires target-hospital validation |
| Signal precedes event recording | A time interval separates signal and recorded event | Compare with recognition proxies first; claims about contemporaneous state additionally require a valid recognizability interval $(t_L^*, t_U^*]$ |
| Inputs strictly satisfy contemporaneous availability | Excludes future inputs and inputs unavailable at query time | Determine whether inputs arose from existing judgment and are available at the intended stage and hospital |
| Care-process features carry predictive information | Prognostic association may include existing clinical judgment | Supports risk stratification within evaluated processes; earlier use requires validation without the relevant information; attribution requires separate evidence |
| Validated future-outcome prediction before a | Predicts the evaluated outcome before the declared proxy; does not | Supports prediction relative to proxy $t_r$; test alternative proxies; current-state recognition |

| Observed evidence | Supported interpretation | Deployment implications and additional evidence |
|---|---|---|
| recognition proxy | establish precedence over actual awareness | additionally requires an as-of reference |
| Performance against an independent departure target | Discrimination and calibration for the defined state-recognition task | A claim of precedence additionally requires using $Z_t$ to validate $p_t$, comparing $t_r$ with $t_{AI}$, and checking $(t^*_L, t^*_U]$; clinical benefit requires separate evaluation |

## Three tests for current-state recognition and their implementation

The tests ask whether departure can be independently defined from contemporaneous evidence, whether the model reliably recognizes that state, and whether thresholded signals precede recognition proxies.

“Independent” means that target definition and assessment are not inferred from the evaluated model or the patient's subsequent care, recognition, or outcome. Reference procedures must suit the target. Requirements for rule-based and manual assessment follow (Fig. 1c). Box 1 lists general and additional reporting items.

Test 1: reference validity. Rules require clinical justification and validation using contemporaneous information; timestamps alone are insufficient. Reviewers see only contemporaneously available records, without subsequent care, final outcomes, or model outputs. Intermix patients with and without subsequent events; absence of the outcome does not imply absence of departure. Prevent hindsight by sequentially revealing records and locking earlier assessments, or assigning different times for one patient to different reviewers.

Prespecify operational criteria and aggregation, including abnormality duration and evidence requirements. Multiple-rater schemes report $\pi(t_k)$, agreement, and original disagreements; rules report validation, indeterminate assessments, and borderline-case handling. Agreement measures reproducibility, while target validity still requires clinical justification and validation.

Test 2: target-matched performance. At patient-query observations with valid references, use $Z_t$ to evaluate discrimination and calibration of state probabilities $p_t$. Report false-alert burden per patient-time at declared thresholds, including counting rules, time denominators, and unassessed periods. Indeterminate states are not automatically negative. Account for within-patient dependence in interval estimates. Self-resolving departures remain transitions; response costs require separate evaluation.

Test 3: threshold crossing relative to recognition. Prespecify the proxy defining $t_r$; repeat analyses with clinically justified, available earlier or later alternatives, explaining absent proxies and routinely triggered orders. Among independently assessed transitions subsequently recognized by that proxy, report the proportion first crossing before $t_r$, interval $t_r - t_{AI}$, and position of $t_{AI}$ relative to $(t^*_L, t^*_U]$.

Conservatively, signals no later than $t^*_L$ are anticipatory predictions; strict ordering within $(t^*_L, t^*_U]$ is indeterminate. Signals after $t^*_U$ but before $t_r$ support recognition after recognizability and before the proxy, with lead time $t_r - t_{AI}$. At the upper bound, recognizability is established, but strict precedence over the signal is uncertain. Report these signals separately, without treating them as absent states or false alerts. This is sufficient, not necessary, for valid state recognition.

After target and performance validation, precedence over the earliest reliable proxy strengthens temporal evidence. Conclusions remain proxy-relative; clinicians' first awareness is unobserved.

Preserve conditions in which the model does not influence care, such as silent operation. Displayed signals may change recognition and treatment; resulting temporal relationships require an appropriate comparative design.

Prespecify patient and query sampling. Programs or trained staff may organize records; clinical interpretation requires qualified personnel. Manual schemes may combine primary review with independent reassessment of a prespecified subsample spanning assessment results, not merely disagreements or alerts. Reassessment size depends on target complexity and precision requirements. Report review volume, time, disagreements, and indeterminate results. Record enrichment-sampling probabilities and use appropriate population weights for performance and absolute burden. Costs and reliability require task-specific validation.

### Box 1. Suggested reporting items matched to capability claims.

#### General items: intended use and prediction under corresponding conditions

1. Specify patients, stage of use, future outcome, horizon, query frequency, outcome ascertainment, and follow-up.

2. Trace key input origins and actual availability; explain relationships to judgment and routine care, and availability at the intended stage.

3. For use before recognition, prespecify observable proxies and eligible windows; report routine orders, absent proxies, and alternative-proxy analyses.

4. Report target-matched discrimination, calibration, and uncertainty; specify population, horizons, sampling distribution, and within-patient dependence.

5. When comparing performance before and after proxies, report changes in samples and horizons; distinguish provenance, sensitivity analyses, and causal attribution.

6. For cross-site use, report target-hospital performance and information conditions; signal or action claims additionally require thresholds and corresponding burdens.

#### Additional items: the state-recognition target and its independence

A1. Specify the departure, clinical basis, and prespecified assessment criteria; explain how current-state assessment avoids inference from model outputs or subsequent care, recognition, and outcomes.

A2. Specify rules or structured review, clinical validity, validation findings, and uncertainty.

A3. Declare query grid, criteria $C(t_k)$, and aggregation; report onset interval $(t^*_L, t^*_U]$, left or right censoring of $T^*$, and observed bounds.

A4. Among recognized transitions, report the proportion without an established recognizable window before $t_r$; retain nonmonotone assessments. Multiple-rater schemes additionally report raw $\pi(t_k)$ and agreement.

#### Information boundaries for as-of state assessment

A5. Specify visible information at each time and preservation of contemporaneous record versions, avoiding backfilling earlier queries with final database values.

A6. Distinguish collection, result generation, and availability; report availability delays.

### Recognition proxies and sensitivity analyses

A7. Declare the proxy defining $t_r$: suspicion, diagnostic assessment, treatment, or documentation.

A8. Repeat analyses with clinically justified alternatives; report temporal findings, absent proxies, and handling of routinely triggered orders.

### Test 1: reference assessment

A9. Reviewers see only contemporaneously available records, intermixing patients with and without subsequent outcomes. Report blinding and qualifications; rule-based references require validation procedures.

A10. Operationalize assessment and aggregation criteria. Report agreement for multiple raters or validation and indeterminate assessments for rules.

A11. Explain prevention of later-record contamination; report sampling, independent reassessment coverage, review volume, and time. Retain raw disagreements and nonmonotone results.

### Test 2: target-matched performance

A12. Against $Z_t$, evaluate state probabilities $p_t$ for distinguishing whether recognizability has been reached.

A13. Evaluate calibration of $p_t$ against assessed states; future-outcome risk is not directly interpretable as current-state probability.

A14. Report false-alert burden per patient-time and counting rules at declared thresholds; do not automatically classify indeterminate states as negative. Account for repeated queries; explain enrichment weights and self-resolving departures.

### Test 3: first-crossing timing

A15. Among independently assessed, subsequently recognized transitions, report the proportion first crossing before $t_r$, interval $t_r - t_{AI}$, and position of $t_{AI}$ relative to $(t_L^*, t_U^*]$.

A16. Classify signals no later than $t_L^*$ as anticipatory predictions; preserve uncertainty within $(t_L^*, t_U^*]$. Signals after $t_U^*$ but before $t_r$ support pre-recognition lead time under a conservative sufficient condition.

A17. Establish absence of model influence on care or describe a design preserving a comparator without displayed signals.

### Additional evaluation when signals trigger action

A18. Specify triggering and response rules; use an appropriate comparative design to evaluate benefits, unnecessary review, and other burdens.

### Information conditions and stages of clinical deployment

Prerequisites must match intended roles. Changes in investigation timing, collection, or recording may alter predictive clues and target relationships. Cross-site deployment requires verifying availability at the intended stage and external validation in the target setting[1]. Multimedia Appendix 1, Section S5, provides a conditional account of this dependence.

Models relying on post-investigation information may lack key inputs before suspicion and test ordering[26]. Such gaps may occur precisely where resources or recognition are limited and support is most needed. Check clue generation, visibility, and availability in the target department; measure actual performance changes. Models primarily supporting post-recognition stratification should be restricted accordingly, rather than directly repurposed before recognition.

Silent operation through the target hospital's real data pipeline can assess latency, missingness, coverage, and stage-specific performance. This tests technical and workflow suitability; benefits and unnecessary review following model-triggered investigation or escalation require separate comparative evaluation[15]. Translate findings into explicit use boundaries and information prerequisites so hospitals can decide when to use the model, which conditions to establish, and which uses lack evidence.

## Conclusion

Predictive performance supports credible deployment expectations only when interpreted alongside clinical stage and information source. We organize evaluation accordingly: specify use, examine input and endpoint generation, and validate performance under corresponding conditions. Current-state recognition claims additionally require an as-of state reference, target-matched performance, and timing relative to recognition proxies. This helps hospitals distinguish supported uses from unvalidated capabilities, avoiding extrapolation of care-process advantages to settings where those clues have not arisen or are unavailable. Before deployment, establish whether the conditions underpinning performance still hold when the model is most needed.

## Acknowledgments

We thank Professor Adrian Weller, Department of Engineering, University of Cambridge; Professor Sheng Dai and Guanli Gu, Department of Medical Affairs, Sir Run Run Shaw Hospital, Zhejiang University School of Medicine; Professor Yan Chen, Department of Infectious Diseases; and Chen Huang, Lili Wu, and Mengjia Zhao, Department of Nursing, for reviewing and supporting this work.

### Use of AI assistance

Codex (OpenAI, accessed October 2026) was used during manuscript development for conceptual and methodological discussion, language revision and translation, reference and mathematical consistency checks, figure-label translation, and document preparation. The authors retain responsibility for all claims, citations, equations, and interpretations.

## Funding

This work received no funding.

## Authors' Contributions

J.Y. developed the framework, wrote the supplementary material, and drafted the manuscript. Z.Z. and L.R.J. supervised the work and critically revised its intellectual content. J.Y., L.R.J., Z.Z., and X.C. approved the final manuscript.

## Conflicts of Interest

None declared.

## Data Availability

Data sharing is not applicable because no datasets were analyzed in this study.

## Abbreviations

AI: artificial intelligence

AUROC: area under the receiver operating characteristic curve

CT: computed tomography

ERM: empirical risk minimization

ICU: intensive care unit

ISGLS: International Study Group of Liver Surgery

ISGPS: International Study Group on Pancreatic Surgery

## Multimedia Appendix 1

Multimedia Appendix 1. Formal Arguments on Clinical Judgment and the Interpretation of Predictive Performance, including Propositions S1–S8 and Corollaries S2.1 and S6.1.

# Formal Arguments on Clinical Judgment and the Interpretation of Predictive Performance

This appendix is Multimedia Appendix 1 referred to in the main text. Clinical judgment can enter model inputs through testing, consultation, monitoring and other clinical activities. These records may contain genuine information about disease and outcomes, so using existing judgment can improve prediction of recorded events. If this performance is interpreted as an ability to identify clinical change before clinical recognition, however, existing judgment may serve as a predictive shortcut. This appendix establishes which conclusions such gains support and which additional evidence is needed concerning the target, timing and origin of the signal.

Sections S1 and S2 establish when actions can convey target information beyond the remaining inputs and how this information reduces optimal predictive risk. Under the stated transmission condition, this gain need not add information beyond the existing clinical assessment. Sections S3 and S4 show, respectively, why event posterior probabilities and lead time relative to an event do not establish current-state recognition or its precedence over clinical recognition. Section S5 gives a conditional mechanism through which differences in care context can affect performance. Section S6 addresses the more detailed validation required by the main text for claims of current-state recognition: a contemporaneous reference, target-matched performance and timing relative to a declared recognition proxy. Evaluating future-outcome prediction by intended stage of use does not require first reconstructing every patient's earliest recognizable time. The arguments use standard probability, information-theoretic and scoring identities, with the assumptions and comparison conditions stated for each result.

Notation. The table summarizes the principal symbols and the section in which each first appears.

| Symbol | Definition | First section |
|---|---|---|
| $B_t$ | The clinician's existing assessment or working hypothesis at time t (denoted B in the static results) | S1 |
| $\mathcal{F}_t$ | The sigma-algebra generated by information actually available by query time t | S1 |
| $X_t = (V_t, A_t)$ | The model's actual inputs: action and workflow features A, together with the remaining inputs V | S1 |
| $T^*$ | The first time the prespecified recognizability criteria are met | S1 |
| $Z_t = 1\{T^* \leq t\}$ | An indicator of whether the prespecified recognizability criteria have ever been met by time t; it remains positive thereafter | S1 |
| $p_t, p_t^*$ | The actual model output for the state probability at time t; the corresponding true conditional probability | S1 |
| $Y$ | A subsequent recorded event (a future or alternative target) | S1 |
| $W$ | A generic binary target in the static results (the current state or a subsequent event) | S1 |
| $h(p)$, $H(W \mid U)$ $I(W; A \mid V)$ | Binary entropy; conditional entropy; conditional mutual information | S1 |

| Symbol | Definition | First section |
|---|---|---|
| $p_0,\ p_1$ | Target probabilities conditional on V and on (V, A), respectively | S2 |
| $m(U),\ L(m),\ R(m),$ | Probability predictor; Logarithmic risk; Brier risk; optimal logarithmic risk using V; information gain δ | S2 |
| $D_{\mathrm{KL}}$ | Kullback–Leibler divergence between Bernoulli distributions | S2 |
| $S = f(V, A)$ | Model output (a probability predictor) | S2 |
| $s(u),\ p(u),\ \alpha,\ \beta$ | Event and state posterior probabilities; coefficients linking the event and state | S3 |
| $D,\ Y(d)$ | Treatment after clinical recognition; the potential outcome under treatment d | S3 |
| $t_{\mathrm{AI}},\ t_r,\ t_E$ | Alert time; recognition-proxy time; recorded event time | S4 |
| $r_g,\ \eta,\ a_g$ | Within-group assessment reliability; probability of an action flip; effective action reliability | S5 |
| $q$ | Case-weighted mean reliability; the output level of the fixed predictor | S5 |
| $\mathrm{Acc}_g,\ R_g$ | Within-group accuracy and Brier risk of the fixed predictor | S5 |
| $w_e,\ p_{\mathrm{src}},\ p_{\mathrm{tgt}}$ | The proportion of cases from group H in environment e; posterior probabilities in the source and target environments | S5 |
| $(t_L^*, t_U^*]$ | An adjudicated interval for the state-transition time, based only on evidence available by the relevant time | S6 |

Proposition locations. Proposition S1 appears in Section S1; Proposition S2, Corollary S2.1, and Proposition S3 in Section S2; Proposition S4 in Section S3; Proposition S5 in Section S4; Proposition S6, Corollary S6.1, and Proposition S7 in Section S5; Proposition S8 in Section S6.

## S1 Existing clinical information can reach a model through recorded actions

Testing, consultation or changes in monitoring can record a clinical assessment formed before the model produces its output. Empirical evidence of the predictive value of clinical action data motivates examination of this input pathway [1]. For actions to convey existing judgment, the assessment must distinguish the target classes, and its recorded expression must retain information not already contained in the model's remaining inputs. The following proposition gives sufficient conditions for this transmission.

Let $B_t$ denote the clinician's existing assessment or working hypothesis, formed by combining prior clinical knowledge with observations of the patient. Clinical concern about a suspected state transition is one expression of this assessment. Prior knowledge here means experience and learned relationships; a Bayesian prior is a probability distribution specified before an explicitly defined update. Neither is equivalent to the updated assessment $B_t$, which is not assumed to be an exact Bayesian posterior distribution. The assessment and its action traces are information-bearing variables, rather than patient-state targets.

Let $\mathcal{F}_t$ be the sigma-algebra generated by information actually available by query time $t$, and let $X_t = (V_t, A_t)$ be the model's actual input, with $\sigma(X_t) \subseteq \mathcal{F}_t$. Here, $A_t$ comprises action-related and workflow features selected from the record and actually supplied to the predictor; $V_t$ contains the remaining inputs. Availability determines what can be selected, whereas selection determines what is supplied. Whether the predictor uses a supplied feature is a separate question addressed in Section S2. This partition isolates one information pathway; it does not assert that measurements or missingness patterns in $V_t$ are independent of clinical behavior. Comparing inputs with and without $A_t$ asks what the action traces add beyond the remaining inputs, even when those traces encode judgments the clinician has already made.

The current-state target of interest corresponds to a recognizable departure from the clinical course, as described in the main text, and is defined by prospectively specified recognizability criteria. If $T^*$ is the first time these criteria are met, set

$$Z_t = \mathbf{1}\{T^* \leq t\}. \quad \text{(S1)}$$

Thus, $T^*$ concerns the first time the recognizability criteria are met, rather than biological onset. The indicator records whether these criteria have ever been met; the historical state remains positive even if the patient subsequently recovers, and does not imply that the abnormality persists. We distinguish $Z_t$ from the subsequent recorded event $Y$. In the static results below, $W \in \{0,1\}$ denotes one fixed target, which may be $Z_t$ or $Y$, with the time subscript suppressed. The relevant results must be applied separately if the target changes.

Consistent with the main text, write $p_t^* = P(Z_t = 1 \mid X_t)$ for the true conditional probability and $p_t = f(X_t)$ for the actual model's state-probability output. The latter estimates the former, and its discrimination and calibration require empirical evaluation. The true conditional probability and the actual predictor are distinguished below for each section's target.

All variables are measurable random elements on a common probability space and take values in standard Borel spaces. Conditional distributions are regular versions, and conditional statements hold almost surely. Within each result, expectations refer to the same specified evaluation population and distribution of query times. These probability conventions are shared across the results; the concern-transmission mechanism, reliability construction and event–state bridge each introduce separate local assumptions. To quantify information about a binary target, write $h(p) = -p\log p - (1-p)\log(1-p)$, use the convention $0\log 0 = 0$ and define:

$$H(W|U) = E h\{P(W = 1|U)\},$$
$$I(W; A|V) = H(W|V) - H(W|V, A). \quad \text{(S2)}$$

Here, $U$ is any specified input variable. Conditional entropy $H$ measures residual uncertainty about $W$; conditional mutual information $I$ measures the reduction in that uncertainty when $A$ is added to $V$. With natural logarithms, information is measured in nats, and $0 \leq I(W; A|V) \leq \log 2$ [2].

Proposition S1. Consider a binary representation in which $W = 1$ denotes the target-positive class, $B = 1$ denotes clinical concern about that target, and $A = 1$ denotes the presence of a specified action-related input feature. The value $A = 0$ means that this feature is absent, rather than that no clinical care was provided. Assume $A \perp W | (B, V)$. This concern-transmission condition describes an information channel in which $B$ and $V$ summarize the target information carried by the action; it is a local mechanistic assumption, not a fact inferred from predictive accuracy. For values of $v$ satisfying $0 < P(W = 1|V = v) < 1$, define:

$$\begin{aligned} b_w(v) &= P(B = 1|W = w, V = v), \\ q_b(v) &= P(A = 1|B = b, V = v). \end{aligned} \quad \text{(S3)}$$

The difference in action probabilities between the target classes satisfies

$$\begin{aligned} &P(A = 1|W = 1, V = v) - P(A = 1|W = 0, V = v) \\ &\quad = [q_1(v) - q_0(v)][b_1(v) - b_0(v)]. \end{aligned} \quad \text{(S4)}$$

If both factors are nonzero on a set of $v$ values with positive probability, then $I(W; A|V) > 0$.

Proof. Conditional independence and the law of total probability give $P(A = 1|W = w, V = v) = q_0(v) + [q_1(v) - q_0(v)]b_w(v)$. Subtracting the expressions for $w = 1$ and $w = 0$ proves equation (S4). A nonzero difference means that the action distributions differ between the two target classes. Both classes have positive conditional probability, and conditional mutual information vanishes only under conditional independence, so the information is strictly positive at these input values. Integrating over a set of positive probability proves the final claim.

Equation (S4) separates the assessment's ability to distinguish the target classes from its expression in the record. The sufficient condition requires both. Nonrandom action alone establishes neither: a routine protocol may not express clinical concern, the assessment may not distinguish the target classes, or the relevant information may already be contained in $V$. If $B$ is a measurable function of $V$, the transmission condition gives $W \perp A|V$ and hence $I(W; A|V) = 0$. Positive information need not be large. Within the proposed assessment-to-action mechanism, equation (S4) explains how existing clinical information reaches the model. The next question is how the prediction objective values this signal.

S2 Predictive gains need not represent information beyond existing judgment

To quantify the value of the added information, hold the target and evaluation population fixed and compare optimal prediction using $V$ alone with that using $(V, A)$. Define the corresponding conditional target probabilities as $p_0 = P(W = 1|V)$ and $p_1 = P(W = 1|V, A)$. Here, $p_0$ is the baseline for a comparison of model inputs, neither the clinician's prior probability nor necessarily a prediction from an earlier time: $V$ may contain all the remaining inputs at the query time. For a measurable probability predictor $m(U) \in [0,1]$, use logarithmic loss and binary Brier loss:

$$\begin{aligned} \ell(w, m) &= - w \log m - (1 - m) \log(1 - m), \\ L(m) = E\ell\{W, m(U)\}&, R(m) = E\{W - m(U)\}^2. \end{aligned} \quad \text{(S5)}$$

A correct prediction made with probability one has zero logarithmic loss; assigning zero probability to the class that occurs gives infinite loss. Logarithmic risk is therefore a nonnegative expectation that may be infinite. Risk is optimized over all measurable probability predictors using the stated inputs. These losses are proper: their expectations are minimized by the true conditional probability [3].

Proposition S2. Let $L_V^*$ and $L_{V,A}^*$ be the infima of logarithmic risk for the two input sets, respectively. Then

$$\begin{aligned} &L_V^* - L_{V,A}^* = I(W; A|V) =: \delta, \\ &R(p_0) - R(p_1) = E(p_1 - p_0)^2, \quad \text{(S6)} \\ &\delta = ED_{KL}\{Bern(p_1) \| Bern(p_0)\}. \end{aligned}$$

Proof. Write $p_U = P(W = 1|U)$, take the conditional expectation given $U$, and add and subtract its binary entropy to obtain:

$L(m) = H(W|U) + ED_{KL}\{Bern(p_U)\|Bern(m(U))\}$.(S7)

The Kullback–Leibler divergence between Bernoulli distributions is $D_{KL}\{Bern(p)\|Bern(m)\} = plog(p/m) + (1-p)log\{(1-p)/(1-m)\}$, with the same endpoint conventions. It is nonnegative and vanishes if and only if $p = m$. The infimum is therefore attained by $p_U$, giving $L^*_V = H(W|V)$ and $L^*_{V,A} = H(W|V, A)$. Their difference is equation (S2). Setting $U = (V, A)$ and $m(U) = p_0$ in equation (S7), and using $L(p_0) = H(W|V)$, gives the expected-divergence identity in equation (S6). For Brier loss, expand $W - p_0 = (W - p_1) + (p_1 - p_0)$. The cross term has zero expectation because $E(W - p_1|V, A) = 0$, proving the second identity.

The information in Proposition S1 has a precise statistical value: the expected change from $p_0$ to $p_1$, measured by KL divergence, equals the reduction in optimal logarithmic risk. This is a conditional link between statistical information and reduction in Bayes risk [4]. The prediction objective values target information without distinguishing whether it comes from clinical judgment already formed. Such gains can therefore be genuine for the declared prediction target while still constituting a shortcut relative to a claim of capability before clinical recognition. Equation (S6) concerns optimal predictors under proper scoring rules; it does not guarantee improvement for every training algorithm or every discrimination metric.

The same risk gap can conditionally establish whether a particular model must use action traces. Let $f(V, A)$ be a fixed measurable probability predictor. The gain $\delta$ is the reduction in optimal logarithmic risk obtained by adding $A$; omitting $A$ forgoes this gain. Let $\varepsilon$ be an allowed upper bound on the predictor's population excess logarithmic risk relative to the full-input optimum. If this bound is smaller than the gain forgone by omitting $A$, a predictor satisfying the bound cannot ignore these inputs entirely.

Corollary S2.1. If $\delta > 0$ and

$L(f) \le L^*_{V,A} + \varepsilon, 0 \le \varepsilon < \delta$,(S8)

then there is no measurable function $k(V)$ such that $f(V, A) = k(V)$ almost surely.

Proof. Such a function would have risk at least $L^*_V = L^*_{V,A} + \delta$, contradicting equation (S8).

This result establishes the predictor's functional dependence on $A$ under a population-risk condition. Model capacity or pretraining scale alone does not establish that condition, and pretraining token loss is different from the clinical-target risk in equation (S8). Functional dependence shows that a predictor uses an input; whether that input conveys clinical concern requires the evidence of origin discussed in Section S6.

The comparison now shifts from the information that $A$ adds beyond the remaining inputs $V$ to the information that the output adds beyond $V$ and the existing assessment $B$. Information gain in the first comparison can be positive while the increment in the second is zero.

Proposition S3. Under the transmission condition $W \perp A|(V, B)$, the output $S = f(V, A)$ satisfies:

$I(W; S|V, B) = 0, I(W; S|V) \le I(W; B|V)$.(S9)

Proof. Given $(V, B)$, the output is a function of $A$, which is conditionally independent of $W$, giving the first equality. Processing $A$ into $S$ cannot increase conditional information, so $I(W; S|V) \leq I(W; A|V)$. Expanding the same joint information in two orders gives:

$$\begin{aligned} I(W; A, B|V) &= I(W; B|V) + I(W; A|V, B) \\ &= I(W; B|V), \\ I(W; A, B|V) &= I(W; A|V) + I(W; B|V, A) \\ &\geq I(W; A|V). \end{aligned} \quad \text{(S10)}$$

Combining these inequalities proves equation (S9), a conditional data-processing argument [2].

Setting $W = Y$ applies the results in Sections S1 and S2 to the same disease or complication endpoint. Under the conditions of Proposition S1, the action traces $A$ supplied as inputs contain information about $Y$ beyond $V$. Proposition S2 quantifies the resulting gain in optimal prediction of $Y$, and Corollary S2.1 gives a population-risk condition under which a particular predictor cannot ignore $A$ entirely. Actions remain inputs, and the disease endpoint is unchanged. Under the transmission condition, Proposition S3 shows that the model output adds no target information beyond the remaining inputs and existing judgment. The positive gain in equation (S6) conditions on $V$, whereas the zero increment in equation (S9) additionally conditions on $B$. Organizing or communicating existing judgment can still be valuable; an improved score does not establish that the model acts before the judgment conveyed by its inputs. For future endpoints, the transmission condition must be examined separately if actions change the outcome or carry patient information not summarized by $B$ and $V$.

S3 A subsequent-event posterior need not identify the current-state transition

An increase in information about $Y$ leaves a further question: what does it establish about the current-state transition $Z_t$? Existing clinical judgment may accurately reflect the patient's state, but an association between the current state and a subsequent event does not make their probabilities interchangeable. To expose the missing link, retain $U = (V, A)$ and write $s(u) = P(Y = 1|U = u)$ and $p(u) = P(Z_t = 1|U = u)$.

Proposition S4. Assume a simple event–state bridge satisfying $Y \perp U | Z_t$, with $\alpha = P(Y = 1|Z_t = 1)$ and $\beta = P(Y = 1|Z_t = 0)$. These coefficients are the probabilities of the subsequent event when the prespecified recognizability criteria have, or have not, been met by time t, respectively. Then:

$$s(u) = \beta + (\alpha - \beta)p(u). \quad \text{(S11)}$$

Under this bridge, if the coefficients are known and $\alpha \neq \beta$, inversion gives:

$$p(u) = \frac{s(u) - \beta}{\alpha - \beta}. \quad \text{(S12)}$$

When the bridge coefficients are unknown, the observable distribution of $(U, Y)$ alone need not identify $p$.

Proof. Conditioning on the two values of $Z_t$ gives $s(u) = \alpha p(u) + \beta\{1 - p(u)\}$, proving equations (S11) and (S12). To show nonidentifiability, fix the same distribution of $U$, with event probabilities $0.3$ and $0.7$ in its two strata. One mechanism has $(\alpha, \beta) = (1,0)$, giving state probabilities $(0.3, 0.7)$. Another has $(\alpha, \beta) = (0.8, 0.2)$, giving state probabilities $(1/6, 5/6)$. In each mechanism, first condition on $U$ to sample $Z_t$, then condition on $Z_t$ to sample $Y$. Both mechanisms define valid joint distributions satisfying the bridge and produce the same observable event probabilities, but different current-state probabilities.

Even exact knowledge of the event posterior cannot distinguish these mechanisms. Resolving the ambiguity requires additional state labels or justified structural information. Known, unequal bridge coefficients provide one sufficient route, but are not required by every possible identification strategy. The bridge is itself a substantive assumption: conditional on the current state, the event distribution must no longer depend on the available inputs.

Care processes create two distinct evaluation problems. When actions express existing concern, the issue is the origin of the input signal. When clinical recognition or care processes affect which events are recorded and when, the issue is the definition and observation of the reference target. These correspond to the input-side and target-side pathways in the main text. Subsequent treatment can also change the future outcome being predicted.

For a discrete treatment variable $D$ after clinical recognition, let $Y(d)$ denote the event outcome under treatment $d$. Consistency states that $Y = Y(D)$ [5]. Partitioning the observed population by treatment actually received gives:

$$P(Y = 1|U) = \sum_d P(Y(d) = 1|U, D = d)P(D = d|U). \quad (S13)$$

Each term remains conditional on treatment actually received. Removing that conditioning requires additional identification assumptions, such as appropriate conditional exchangeability; consistency alone does not make this observed mixture an untreated risk or a risk under a specified intervention. Treatment after clinical recognition can change $Y$, so the observed event and the current state answer different questions. A current-state reference avoids defining the present directly by the outcome of subsequent treatment, although prior treatment can still affect the patient's course. Using a future-event label as a reference for a current-state transition therefore requires separate justification. Once the reference target is established, the ordering of the model output and clinical recognition still requires separate validation.

S4 Better event prediction can coexist with alerts after clinical recognition

For a threshold-based system, let $t_{AI}$ be the first alert time under the declared threshold and $t_r$ a prespecified, observable proxy time for clinical recognition. The alert time is not the first time predictive information becomes available, and the proxy does not directly observe the onset of the clinician's internal concern. Action features in $A$ do not automatically serve as the clinical-recognition reference, although the two uses may coincide if explicitly declared. For patients with a subsequent recorded event, let $t_E$ be its recorded time. All times use the same clock. Equation (S14) applies to individuals for whom all three times are observed and finite; those with no alert, a missing recognition proxy or no event during follow-up must be reported separately as non-occurrence, missingness or censoring, as appropriate. The following decomposition underlies the evaluation of timing relative to clinical recognition in Test 3 of the main text.

Proposition S5. Lead time relative to the event and lead time relative to clinical recognition satisfy:

$$t_E - t_{AI} = (t_r - t_{AI}) + (t_E - t_r). \quad (S14)$$

Perfect event discrimination and an arbitrarily long positive lead time relative to the event can coexist with alerts issued only after the recognition-proxy time.

Proof. Adding and subtracting $t_r$ gives equation (S14). For the second claim, take $Y$ to be an independently defined disease or complication endpoint with balanced positive and negative cases, and let $V$ be constant. At time $1$, the clinician's concern perfectly distinguishes this subsequent endpoint, so $B = Y$, and generates the action input $A = B$. Thus, $A = Y$ in this construction equates their binary values without making them the same clinical object: disease remains the prediction target, and action is only an input. At time $2$, the model reads this action and outputs $S = A$, issuing an alert when $S = 1$. For event-positive patients, set $t_r = 1$ and $t_E = 2 + d$, where $d > 0$. All positive outcomes receive score $1$ and all negative outcomes score $0$, giving event AUROC $1$. The two lead times, however, are:

$$t_E - t_{AI} = d, t_r - t_{AI} = -1. \quad \text{(S15)}$$

This construction refutes a general inference from event-prediction performance and lead time relative to the event to capability before clinical recognition.

This fixed-endpoint construction also illustrates the scoring and transmission results in Section S2. With $V$ constant and $A$ withheld, the optimal disease probability is $1/2$, Brier risk is $1/4$ and logarithmic risk is $\log 2$. Once $A$ is supplied, $S = A$ achieves zero risk under both losses for the same disease endpoint. The information gain is $I(Y; A|V) = \log 2$, whereas $I(Y; S|V, B) = 0$: the information improving prediction was already present in clinical concern. The model reads an accurate, genuinely available signal after the judgment expressed by that signal has formed. The event time can be an independently verified disease-event time recorded without additional delay; this input-side mechanism requires neither shifting the event time nor redefining its label. Nor does it require temporal or availability leakage, which would expose information not yet available at the query time. Perfect clinical concern is used here to construct a simple logical counterexample, not to characterize typical clinical accuracy.

Equation (S14) also identifies the component affected by delayed event recording. When $t_E \geq t_r$, the interval from the recognition proxy to the event increases apparent lead time without being created by the model. Delaying only $t_E$ by $d$ increases reported event lead time by $d$ while leaving timing relative to clinical recognition unchanged. This comparison holds the patients, labels and model outputs fixed; reselecting patients as the event time changes would introduce a separate selection effect. A long event lead time therefore cannot replace a valid current-state reference or evidence that the alert precedes the declared recognition proxy.

S5 Reliance on clinically conveyed information can make performance depend on care context

The predictive value of an action may vary with the task-specific reliability of the clinical assessment it expresses. To isolate this mechanism, the following construction holds the model and target distribution fixed while varying only task-specific assessment reliability. It gives a conditional explanation for performance differences across care contexts, without ranking clinicians by seniority, institution type or overall competence. This context-dependence result is separate from the preceding target and timing requirements.

Set $W = Z_t$, let $V$ be constant, and assume that each clinician group $g$ encounters the same balanced target distribution, $P(W = 1|g) = 1/2$. Target probabilities before observing the assessment are therefore unchanged across groups; what varies is the likelihood of the assessment given the target state, rather than the prior target probability. The model does not receive group membership. The binary assessment expressed as clinical concern has symmetric, task-specific reliability $r_g \in (1/2, 1]$:

$$P(B = 1|W = 1, g) = P(B = 0|W = 0, g) = r_g. \quad \text{(S16)}$$

The action expresses clinical concern through $A = B \oplus N$, where $\oplus$ denotes binary exclusive OR, $N \sim Bern(\eta)$ is independent of $(W, B, g)$, and the noise parameter shared by all groups satisfies $0 \leq \eta < 1/2$. The action is correct when correct concern is not flipped or incorrect concern is flipped. Its effective reliability is therefore:

$$a_g = P(A = W|g) = r_g(1 - \eta) + (1 - r_g)\eta \\ = \eta + (1 - 2\eta)r_g. \quad \text{(S17)}$$

Making $V$ constant isolates this information channel without leaving unmodeled interactions with other inputs. This is stronger than assuming only that $V$ is marginally uninformative: information can reside in combinations of components even when a component alone predicts nothing.

Proposition S6. Let $q = E_{train} a_g > 1/2$, where the expectation uses the case-weighted group mixture in the training population. The Bayes probability predictor under the training distribution is:

$$f(A) = \begin{cases} q, & A = 1, \\ 1 - q, & A = 0. \end{cases} \quad \text{(S18)}$$

With this predictor fixed, $Acc_g$ denotes its classification accuracy in group $g$ at threshold $1/2$, and $R_g$ its expected Brier loss conditional on that group. Both evaluate the same fixed $f$, rather than predictors reoptimized for each group. They are:

$$Acc_g = a_g, \\ R_g = a_g(1 - q)^2 + (1 - a_g)q^2 \\ = q^2 + (1 - 2q)a_g. \quad \text{(S19)}$$

Consequently, with $q$ and $\eta$ held fixed:

$$\frac{\partial R_g}{\partial r_g} = (1 - 2q)(1 - 2\eta) < 0. \quad \text{(S20)}$$

Proof. Symmetry gives $P(A = 1|g) = 1/2$ and $P(W = 1, A = 1|g) = a_g/2$. Averaging over the training-case mixture and applying Bayes' rule gives $P(W = 1|A = 1) = q$ and $P(W = 1|A = 0) = 1 - q$, proving equation (S18). Because $q > 1/2$, the thresholded prediction equals $A$. When $A = W$, its squared error is $(1 - q)^2$, and otherwise it is $q^2$, giving equation (S19). Substitution of equation (S17) and differentiation prove equation (S20).

The model's reliance on action has thus been derived from the training distribution before comparing group performance. For an exact example, take $\eta = 0$, $r_H = 0.9$ and $r_L = 0.6$, with equal case weights for the two groups in training. The fixed predictor has $q = 0.75$.

| **Quantity in the constructed example** | **Higher-reliability group** | **Lower-reliability group** |
|---|---|---|
| Predicted probability when the action is present | 0.75 | 0.75 |
| True target probability when the action is present | 0.90 | 0.60 |
| Classification accuracy | 0.90 | 0.60 |
| Brier risk | 0.1125 | 0.2625 |
| Absolute calibration error at either output value | 0.15 | 0.15 |

When the action is present (A = 1), the same predicted probability is too low in the higher-reliability group and too high in the lower-reliability group. Accuracy and Brier risk are both better in the group with the more reliable clinical signal, but the absolute calibration error is identical. Brier risk in the lower-reliability group even exceeds the risk of the constant prediction $0.5$, which is $0.25$: the fixed model assigns more confidence to the action than its local reliability supports. These are analytical values, not observed clinical effect sizes. Under the stated assumptions, they establish a possible mechanism and an ordering under specific metrics, rather than an ordering of arbitrary models under every metric.

Corollary S6.1. Keep the predictor and both within-group mechanisms fixed, and let $w_e$ denote the proportion of cases in environment $e$ that come from group $H$. Then:

$$\begin{aligned} a_e &= w_e a_H + (1 - w_e) a_L, \\ Acc_e &= a_e,\ R_e = q^2 + (1 - 2q) a_e. \end{aligned} \tag{S21}$$

For the numerical construction above:

$$Acc_e = 0.6 + 0.3 w_e,\ R_e = 0.2625 - 0.15 w_e. \tag{S22}$$

Proof. Condition on group membership and average the fixed within-group accuracies and risks in equation (S19).

A change in case composition can therefore translate a clinician-level mechanism into performance differences between environments. In this construction, $P(A = 1|e) = 1/2$ remains unchanged while $P(W = 1|A = 1, e) = a_e$ changes. Monitoring only the marginal frequency of the action would miss this change in its predictive meaning.

Beyond this particular mixture, the relevant statistical quantity is the target probability conditional on the model input. Let $U = (V, A)$ and define the source- and target-environment posteriors as $p_{\mathrm{src}}(u) = P_{\mathrm{src}}(W = 1|U = u)$ and $p_{\mathrm{tgt}}(u) = P_{\mathrm{tgt}}(W = 1|U = u)$, respectively. The clinical query time is fixed in this comparison: the subscripts src and tgt denote the source and target environments, rather than different clinical times. Assume $P^U_{\mathrm{tgt}} \ll P^U_{\mathrm{src}}$: any input set with zero probability in the source environment also has zero probability in the target environment. A posterior version determined almost everywhere under the source distribution is therefore unambiguous under the target distribution as well.

Proposition S7. When evaluated in the target environment, the fixed source-environment posterior $p_{\mathrm{src}}$ has the following excess risk relative to the target-environment optimal predictor:

$$\begin{aligned} L_{\mathrm{tgt}}(p_{\mathrm{src}}) - L_{\mathrm{tgt}}(p_{\mathrm{tgt}}) &= E_{\mathrm{tgt}} D_{KL}\{Bern(p_{\mathrm{tgt}}(U)) \| Bern(p_{\mathrm{src}}(U))\}, \\ R_{\mathrm{tgt}}(p_{\mathrm{src}}) - R_{\mathrm{tgt}}(p_{\mathrm{tgt}}) &= E_{\mathrm{tgt}}\{p_{\mathrm{src}}(U) - p_{\mathrm{tgt}}(U)\}^2. \end{aligned} \tag{S23}$$

Proof. Apply equation (S7) under the target distribution to the predictor $p_{\mathrm{src}}$ and subtract the finite optimal logarithmic risk $H_{\mathrm{tgt}}(W|U)$. For Brier loss, expand $W - p_{\mathrm{src}} = (W - p_{\mathrm{tgt}}) + (p_{\mathrm{tgt}} - p_{\mathrm{src}})$ and use $E_{\mathrm{tgt}}(W - p_{\mathrm{tgt}}|U) = 0$.

If the posteriors differ on a set of positive probability under the target-environment distribution, the excess risk is strictly positive and may be infinite for logarithmic loss. The comparator is the optimal predictor in the target environment, not the absolute risk in the source environment. As the preceding construction shows, clinician composition or the translation of concern into action can change this posterior; identifying their contributions in a particular institution requires data on those mechanisms.

Note S5.1 (information ordering under separate optimization). A complementary question is whether one signal can support better prediction than another when each predictor is optimized separately. This extension is not a premise of the fixed-model construction: it allows nonconstant $V$ and compares signal information under separate optimization. Under a common joint distribution, assume $W \perp A_L | (V, A_H)$, so the less informative signal $A_L$ can be generated from $(V, A_H)$ through a conditional random channel. Use the same standard Borel decision space, a common measurable loss bounded below and all measurable randomized decision rules. Any rule based on $(V, A_L)$ can be reproduced from $(V, A_H)$ by first simulating the channel and then applying that rule (Blackwell randomization [6]). The resulting joint distribution of the target, remaining inputs and decision is identical. Taking infima therefore shows that optimal risk using $(V, A_H)$ is no greater than optimal risk using the other signal. For logarithmic loss, the exact difference is:

$$H(W|V, A_L) - H(W|V, A_H) = I(W; A_H|V, A_L) \geq 0, \quad \text{(S24)}$$

because $H(W|V, A_H, A_L) = H(W|V, A_H)$. This channel ordering is stronger than an ordering of clinicians' overall accuracy, and does not order the risks of an arbitrary common fixed predictor.

These comparisons show how the predictive value of clinically conveyed information depends on the context in which it is generated and evaluated. They provide conditional mechanisms, rather than proving that transfer must degrade a model's performance. The next section returns to the capability claim of interest: which reference and timing evidence is needed, and which questions about signal origin remain unresolved.

## S6 Contemporaneous reference and timing evidence for claims of current-state recognition

The preceding results distinguish genuine predictive information from interpretations of capability based on that information. Sections S1 and S2 address information conveyed by existing judgment, Section S3 the current-state reference, Section S4 timing relative to clinical recognition and Section S5 context dependence. For claims that the model has identified clinical change at that time and has done so before the recognition proxy, this section connects the three detailed tests in the main text. Proposition S8 gives a conservative sufficient timing condition, conditional on validity of the adjudicated interval. These tests do not replace all evaluation of future-outcome prediction, nor do they identify the clinical origin of every predictive signal.

Let $(t_L^*, t_U^*]$ be an adjudicated interval for $T^*$ based only on evidence available by the relevant time. Construct this interval for the query or alert being evaluated, using prespecified recognizability criteria and the information available at that time. Validity means that $t_L^* < T^* \leq t_U^*$. The reference procedure should not define the current state by consulting future clinical recognition, future outcomes or the output of the model under evaluation; its evidence criteria and adjudication safeguards remain part of establishing interval validity.

Proposition S8. Conditional on interval validity, the conservative criterion

$$t_U^* < t_{AI} < t_r \quad \text{(S25)}$$

implies $T^* < t_{AI} < t_r$.

Proof. Validity gives $T^* \leq t_U^*$. Combining this inequality with equation (S25) yields the stated temporal ordering.

This ordering places the alert after the state transition has become prospectively recognizable and before the declared recognition proxy. Requiring the alert to fall strictly after the upper bound is sufficient but not necessary: when $t_{AI} = t_U^*$, a valid interval still implies $T^* \leq t_{AI}$, so the state is already positive at that time, but the transition need not strictly precede the model time. Conversely, $t_{AI} \leq t_L^*$ establishes that the alert precedes the transition, whereas $t_L^* < t_{AI} < t_U^*$ leaves their ordering unresolved. Following the reporting categories in Test 3 of the main text, alerts at or before the lower bound are prospective predictions; for alerts within the open interval, whether the state has already occurred remains uncertain. Alerts exactly at the upper bound are reported separately: the uncertainty concerns strict temporal precedence only and is not a basis for treating the state as absent or the alert as false. Only alerts strictly after the upper bound and before the recognition proxy meet the conservative criterion in equation (S25). This algebra makes the interval interpretation precise, but does not establish the validity of the adjudicated interval or the recognition proxy.

Establishing the target and temporal ordering does not resolve the question of origin. Attributing a model's predictive signal to existing clinical concern requires evidence about the origin of its action inputs; even for a transparent predictor, reliance on those inputs does not identify their clinical origin. Take $V$ to be constant, let $W \sim Bern(1/2)$, and suppose that $A = W$ and $S = A$ are observed. One mechanism has $B = W$ and $A = B$, so the action transmits clinical concern perfectly. In another, $B$ is independent of $W$, and a routine process triggered by the state directly generates $A = W$. The mechanisms produce the same distribution of the observed variables $(W, V, A, S)$ and use the same transparent predictor, but only the first contains a concern-mediated pathway. The second mechanism deliberately violates the transmission assumption in Proposition S1. Without information distinguishing the mechanisms, reliance on action cannot identify whether that assumption holds clinically.

The three tests serve distinct evidentiary roles. Test 1 requires an independently specified state reference based only on information available at the time, together with evidence of its validity. Test 2 evaluates discrimination, probability calibration and false-alert burden against that reference. Test 3 evaluates alert timing using valid temporal bounds and a clinically justified recognition proxy. A single alert falling within the required interval does not establish model performance. A reference target that does not depend on subsequent recognition, outcomes or model outputs does not make the model's information statistically or causally independent of clinical judgment. Even when restricted to information available at the time, the state reference may use observations collected because of clinical concern. Timestamped concern, reasons for taking action and sensitivity analyses using alternative proxies can provide further evidence concerning origin.

These formal results show that even genuine predictive information can support an interpretation of clinical capability that exceeds the evidence. Under the transmission assumption, scoring gains need not add information beyond the existing assessment; even an alert after clinical recognition can have positive lead time relative to the event. Conditional constructions further show that changes in the generation or reliability of action signals can affect performance across environments. Future-outcome prediction should be evaluated at the relevant stage of use and under the corresponding information conditions. Claims of current-state recognition and precedence over a clinical-recognition proxy require a valid state reference, target-matched performance and timing evidence. Each type of evidence supports a bounded conclusion; signal origin, applicability across settings and actual patient benefit require designs addressing those respective questions.